\documentclass[letterpaper, 10 pt, conference]{ieeeconf}  % Comment this line out if you need a4paper

\IEEEoverridecommandlockouts                              % This command is only needed if 
\title{Graph-based Cluttered Scene Generation and Interactive Exploration \\using Deep Reinforcement Learning}

\usepackage{mathrsfs}
\usepackage{amsmath}
\usepackage{amsfonts}
\usepackage[table,xcdraw,usenames,dvipsnames]{xcolor}
\usepackage{wrapfig}
\usepackage{soul}
\usepackage{longtable}
\usepackage{subcaption}
\usepackage{mwe}
\usepackage{multirow}
\newcommand{\niranjan}[1]{\textcolor{TealBlue}{{}}}
\newcommand{\sehoon}[1]{\textcolor{magenta}{{}}}

\newcommand{\irfan}[1]{\textcolor{red}{{}}}

\long\def\ignorethis#1{}

\newcommand{\etal}{{\em{et~al.}\ }}

\newcommand{\ie}{i.e.\ }

\newcommand{\vc}[1]{\ensuremath{\boldsymbol{#1}}}

\makeatletter
\let\oldlt\longtable
\let\endoldlt\endlongtable
\def\longtable{\@ifnextchar[\longtable@i \longtable@ii}
\def\longtable@i[#1]{\begin{figure}[t]
\onecolumn
\begin{minipage}{0.5\textwidth}
\oldlt[#1]
}
\def\longtable@ii{\begin{figure}[t]
\onecolumn
\begin{minipage}{0.5\textwidth}
\oldlt
}
\def\endlongtable{\endoldlt
\end{minipage}
\twocolumn
\end{figure}}
\makeatother

\usepackage{longtable}
\usepackage{booktabs}
\usepackage[final]{changes} % Comment to see changes
\usepackage{hyperref}
\usepackage{url}

\author{\authorblockN{K. Niranjan Kumar}
\authorblockA{Georgia Institute of Technology}
\and
\authorblockN{Irfan Essa}
\authorblockA{Georgia Institute of Technology}
\and
\authorblockN{Sehoon Ha}
\authorblockA{Georgia Institute of Technology}}

\begin{document}
\maketitle
\thispagestyle{empty}
\pagestyle{empty}

%===============================================================================
\begin{abstract}
We introduce a novel method to teach a robotic agent to interactively explore cluttered yet structured scenes, such as kitchen pantries and grocery shelves, by leveraging the physical plausibility of the scene. We propose a novel learning framework to train an effective scene exploration policy to discover hidden objects with minimal interactions.
First, we define a novel scene grammar to represent structured clutter. Then we train a Graph Neural Network (GNN) based \emph{Scene Generation} agent using deep reinforcement learning (deep RL), to manipulate this \emph{Scene Grammar} to create a diverse set of stable scenes, each containing multiple hidden objects.
Given such cluttered scenes, we then train a \emph{Scene Exploration} agent, using deep RL, to uncover hidden objects by interactively rearranging the scene. 
% We show that the learned agents can construct cluttered scenes with more hidden objects and efficiently discover hidden objects in complex scenes with higher success rates than the baselines. 
We show that our learned agents hide and discover significantly more objects than the baselines. We present quantitative results that prove the generalization capabilities of our agents.
% and also generalize to scene sizes unseen during training.
% We demonstrate the generalization capabilities of our agents and provide quantitative comparisons that show the effectiveness of our approach over baseline methods.
We also demonstrate sim-to-real transfer by successfully deploying the learned policy on a real UR10 robot to explore real-world cluttered scenes. The supplemental video can be found at: \href{https://www.youtube.com/watch?v=T2Jo7wwaXss}{https://www.youtube.com/watch?v=T2Jo7wwaXss}.%\sehoon{which robot?}
\end{abstract}

\section{Introduction}
%Why is do we need interactive exploration
% A long standing goal of robotics has been to build household robots that can help with everyday chores like cooking or cleaning. Dealing with such challenging real world scenarios requires an agent to actively interact with and explore the environment around it. Consider the following scenario: You get off work and you want to buy groceries on your way home. But you don't remember what you have in your fridge or pantry. You ask your robotic household assistant, what groceries you have at home. To answer such a question, the robot should know a priori, what objects are present in your kitchen. When you were away from home, if the robot had explored it's environment and had an internal representation of the kitchen, it can quickly respond from this knowledge. State-of-the-art object detectors can detect objects in a scene accurately, sometimes even when they are partially occluded. But a typical kitchen has dozens of products all stacked away in small pantry cupboards and fridges. Using a passive object detection framework is insufficient in such cluttered environments with heavily occluded objects. In these environments the robot has to \textbf{interact} with the scene, gather additional information to estimate the true underlying state of the scene, i.e. what objects are present and where they are located.

A long-standing goal of robotics is to build household robots that help with everyday chores like cooking or cleaning. Such tasks involve challenging real-world scenarios, which require a clear understanding of the environment. Computer Vision (CV) is a useful sensory modality to understand such scenes. Nevertheless, it only provides limited information when the scene is cluttered with objects, with many objects partially or completely occluded. For instance, a typical kitchen has dozens of products, all stacked away in small pantry cupboards and fridges. To estimate the underlying state of such a scene, a robotic agent might rely on interactions via manipulators, to augment its visual perception.
% Developing such an efficient agent that can fully understand the scene with minimal interaction is not a straightforward problem in scenes with heavy visual occlusions resulting from dense clutter.
% An intelligent robotic agent thus requires an additional approach to estimate the underlying state of these scenes, such as interactions via manipulators. 
However, developing an efficient agent that fully understands a scene, with minimal interaction, is not a straightforward problem, especially in scenes rampant with heavy visual occlusions and dense clutter.

\begin{figure}[t]
\centering
\begin{subfigure}{0.32\linewidth}
\includegraphics[width=0.95\linewidth,height=0.6\linewidth]{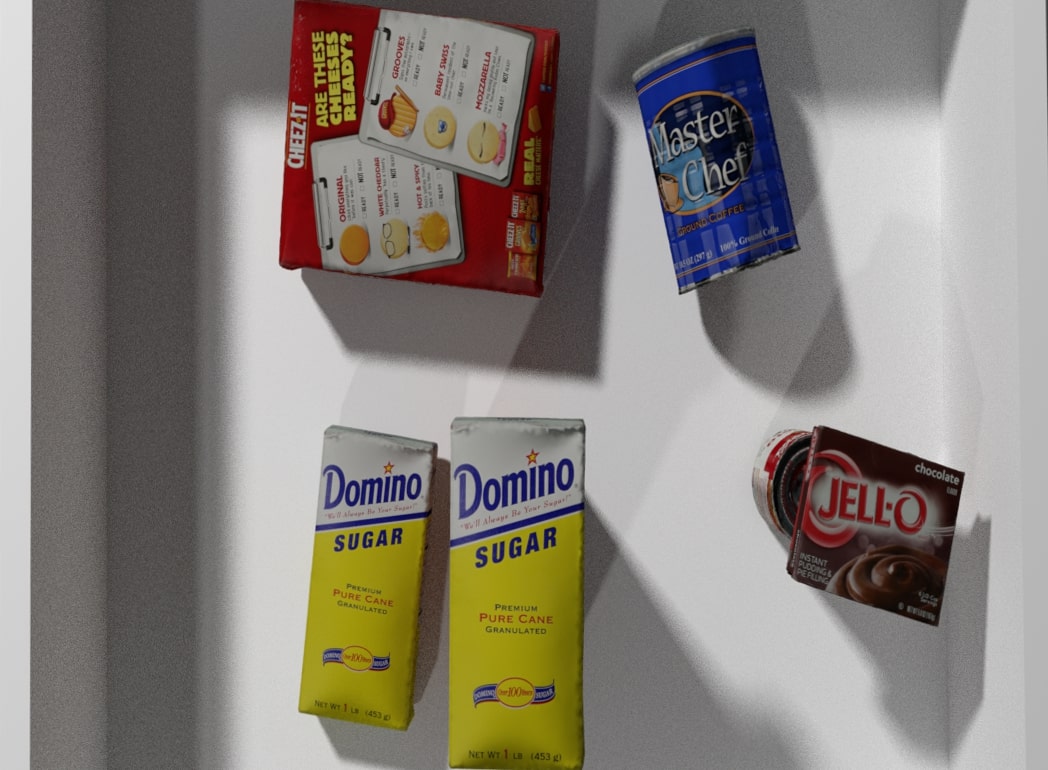}
\caption{Top-view}
\label{fig:cluttered_scene}
\end{subfigure}%
\begin{subfigure}{0.32\linewidth}
\includegraphics[width=0.95\linewidth,height=0.6\linewidth]{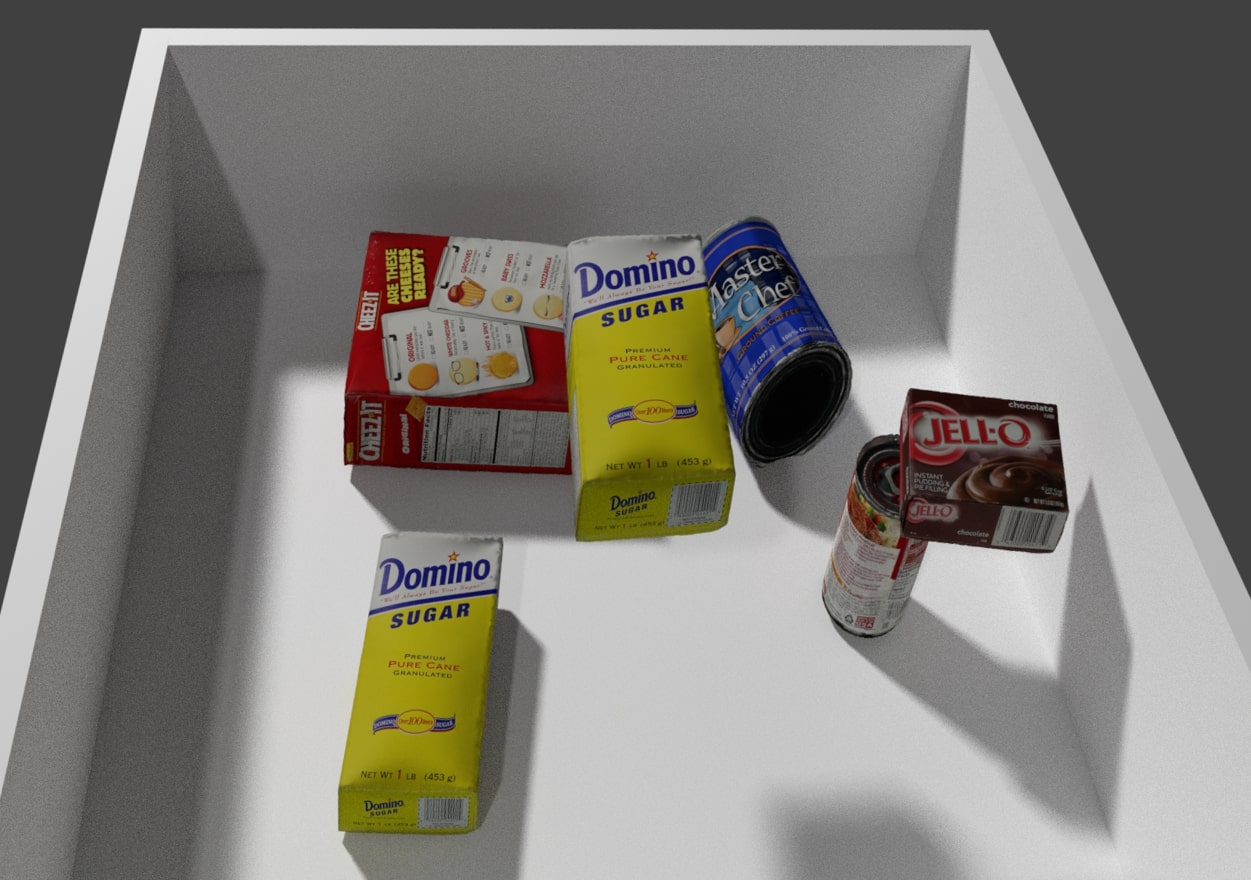}
\caption{Implausible}
\label{fig:cluttered_scene_expectation}
\end{subfigure}%
\begin{subfigure}{0.32\linewidth}
\includegraphics[width=0.95\linewidth,height=0.6\linewidth]{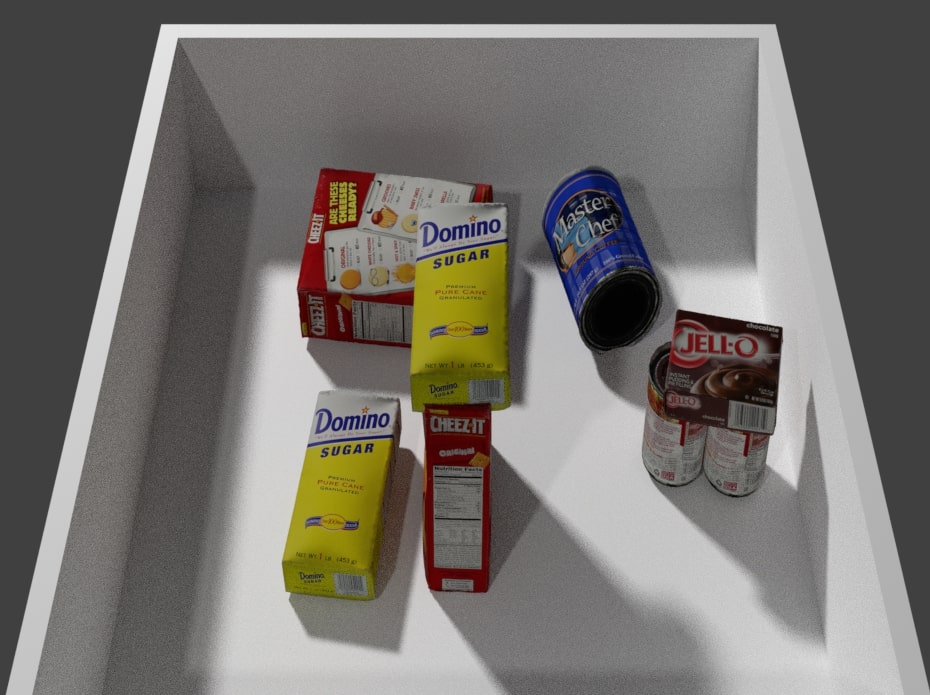}
\caption{Plausible}
\label{fig:cluttered_scene_reality}
\end{subfigure}%
\label{fig:motivating images}
\caption{We show how the agent can leverage clues about physical plausibility to search for objects better. When encountering the scene in Figure \ref{fig:cluttered_scene}, an agent equipped with a static camera cannot completely observe it. 
However, from just the top-view, a knowledge about physics priors can hint that scene (b) is impossible. The agent can then plan smarter interactions to reveal the true state of the scene (c).
% This can lead to physically implausible belief about the world (Figure \ref{fig:cluttered_scene_expectation}) when the actual scene state requires (Figure \ref{fig:cluttered_scene_reality}) requires further interactive exploration.
}
\vspace{-5mm}
% The robot can use such clues to determine which object it should look under to find the hidden objects in the scene.\sehoon{Can you label each figure, like (a) top-view observation (b) physically impossible (c) physically plausible} \sehoon{The caption can be shortened.}}
\end{figure}

% \realworldclutter
The key for developing such a robotic agent is to \emph{connect} visual observations to physics priors. Let us consider Figure~\ref{fig:cluttered_scene}, that shows a cluttered scene of kitchen objects. While we cannot see the entire scene from the top, we can strategically explore it by, for instance, looking under the largest object as it is the most likely to be occluding another object. However, an intelligent agent should utilize additional clues of physical plausibility. 
For example, it can realize that some configurations (Figure~\ref{fig:cluttered_scene_expectation}) are not physically plausible and that there is likely to be an object underneath (Figure~\ref{fig:cluttered_scene_reality}), even though these objects are not visible. Indeed, humans have this intuitive physics engine built into our cognition~\cite{spelke2007core} that helps us notice violations of physical laws, thereby aiding us in exploration. However, building this intuition in a machine, by explicitly defining a set of rules is tedious and impractical. We instead want the agent to \emph{learn} these priors implicitly from a large number of experiences.%\irfan{Motivation is too long. And unclear as to how we actually solve the hard problems mentioned so far!}\niranjan{Any thoughts on what can be removed?}
% looks at a scene from the top and notices that an object is floating in space as shown in Figure \ref{fig:cluttered_scene_expectation}, it can realize that there should be another object under it supporting its weight. If it sees an object stacked on top of another object, but realizes that the arrangement is inherently unstable then it can infer that the scene is not physically plausible unless there's another invisible object supporting it. We humans have this intuitive physics engine built into our cognition \cite{spelke2007core} that helps us notice violations of physical laws in scenes which aid us in exploration. 

In this work, we propose a novel framework for learning an effective strategy to explore cluttered environments by (1) training \emph{Scene Generation} and \emph{Scene Exploration} agents, (2) that utilize a shared graphical representation of the scene, built upon a \emph{Scene Grammar}. First, we define the Scene Grammar to represent the space of possible structured clutter scenes compactly. In this space of possible scenes afforded by the grammar, a majority would either be physically unstable or would not contain any hidden objects. To reliably generate stable scenes with multiple hidden objects, we train a Scene Generation Agent using deep reinforcement learning (deep RL). Then, we train an efficient Scene Exploration agent that can explore these challenging scenes generated with minimal number of interactions. For efficient training, we adopt the ``learning by cheating'' framework~\cite{chen2020learning} that first trains a teacher agent with privileged information and then clones the learned behaviors to a student agent. We model our agents as Message Passing Neural Networks (MPNNs)~\cite{zhou2020graph} and train them with RL to maximize their respective objectives.
% We represent cluttered scenes as \emph{scene graphs} and develop a system of \emph{graph grammar} that can be used to generate a wide range of cluttered scenes. 
% In this space of possible scenes afforded by the grammar, a majority would either be physically unstable or not contain any hidden objects. 
% To reliably generate stable scenes, with multiple hidden objects, we train a Scene Generation Agent using deep reinforcement learning (deep RL).
% These scenes offer a challenging training ground to learn an efficient Scene Exploration Agent.
% Given such scenes, we train a Scene Exploration Agent to interact and decipher the underlying scene graph by leveraging learned clues about physical plausibility. Our method uses CAD models of the objects to simulate cluttered scenes during training. However, during testing, we only need the class ID and 6D poses of the detected objects, which can be obtained from an off-the-shelf pose detector such as PoseRBPF~\cite{deng2021poserbpf}. We model our agents as Message Passing Neural Networks (MPNNs) and train them with RL to maximize their respective objectives. 

We demonstrate that our framework trains an effective Scene Exploration agent that explores complex scenes generated by the Scene Generation agent.
We show quantitative comparisons that indicate significant improvements over baselines and provide evidence of sim-to-real transfer, by testing our method trained in simulation on real world clutter.
% \sehoon{It is very odd that there's something behind the contributions. }
% \sehoon{Too long paragraph! Split into the summary of the approach and the summary of the results.}\niranjan{fixed}

The contributions of our work are as follows: 
% \sehoon{Can you condense the contributions to one concise sentence?} \irfan{Wouldn't that just be something already stated above "leverage learned priors  about  physical  plausibility  to  explore  scenes  with structured  clutter." .. maybe worth stating here again.  I like a list of contributions, but these don't need to be as verbose as they are now}
% \sehoon{I meant, one for each :) I discussed this with Niranjan already.}

% \niranjan{We present a graph based framework to represent, generate and explore complex structured clutter scenes while ensuring physical stability.}
\begin{enumerate}
    \item A scene grammar to generate scene graphs of various cluttered multi-object scenes.
    \item A learning framework to train an effective scene exploration agent that discovers hidden objects, along with a scene generation agent that specializes in hiding objects in clutter.
    \item Demonstration of sim-to-real transfer of our learned exploration policy.
    % which 
    % These scene graphs can be used to synthesize \replaced{a wide range of cluttered scenes and}{challenging and physically plausible scenes, and further} 
    % serve as the backbone representation of our interactive perception framework.
    % \item We represent the state of cluttered multi-object scenes using a scene graph that captures the underlying structure of the scene. We propose a scene grammar that can be used to synthesize and express such scene graphs use them as the backbone representation of our interactive perception framework.
    % \item A Scene Generation Agent that procedurally applies grammar rules to create stable cluttered scenes that maximize the prediction error of a passive perception system.
    % \replaced{These scenes occupy subspace within the space of all possible scenes that can be generated by the scene grammar and are chosen such that they}{The scenes are generated with an adversarial intent to} maximize the prediction error of a passive perception system.
    % \item A Scene Exploration Agent that can locate and identify all the objects present in the scene by taking a series of rearrangement actions that can efficiently uncover previously hidden objects.
\end{enumerate}

%===============================================================================
\section{Related Work}
Our work falls within the broad domain of interactive perception~\cite{bohg2017interactive}, which involves leveraging interaction to infer otherwise hidden properties of an environment, such as masses of objects~\cite{xu2019densephysnet,kumar2019estimating}, state of a cluttered scene~\cite{novkovic2020object,poon2019probabilistic} or kinematic structure of articulated objects~\cite{katz2011interactive}.
% We break down the prior work into three categories: object search, scene generation, and object rearrangement.
% More specifically, our work share common ground with object search, scene generation and object rearrangement, which we discuss in the following subsections.
\paragraph*{Object Search in Robotics}
Building robots that help in domestic households is a long-standing goal of robotics. It is important for such robots to search and retrieve objects in cluttered scenes. 
% \deleted{This involves first finding an object, singulating it, planning a grasp, and then actually executing motor commands. Let us assume that modern CNN-based approaches are powerful enough to find objects despite significant occlusions.} 
However, even when all objects are completely visible, retrieving is quite challenging if many objects are in close proximity. Therefore, a body of work in object singulation focuses on moving objects in clutter such that they are spread out and do not touch each other~\cite{sarantopoulos2019split, chang2012interactive,eitel2017learning}. Once the objects are singulated, planning of feasible grasps becomes much easier as grasping does not depend on the clutter surrounding it. 

Other approaches deal with occluded objects that may be partially or completely invisible~\cite{kurenkov2020visuomotor, danielczuk2019mechanical, zeng2018learning, deng2019deep, yang2020deep}. Often in such heavily cluttered scenes, such as a bin~\cite{danielczuk2020x,kurenkov2020visuomotor,novkovic2020object,danielczuk2019mechanical} or a shelf~\cite{li2016act,huang2020mechanical}, the agent has to move other objects around to search for the target.
% \deleted{While learning to retrieve a target object from a collection of objects thrown randomly into a bin is useful in some scenarios such as warehouses, structured clutter is more common in spaces with where humans interact, like kitchen pantries or supermarket shelves. Arranging objects with some underlying structure helps us recollect where each object is located, aiding in quick retrieval. Additionally, it also ensures that the objects themselves do not get damaged due to large localized contact forces, frequently encountered within object piles in bins.}
% Another distinction that has been made by prior work is on the nature of clutter itself rather than its location. Objects can either be randomly dropped into a bin\cite{danielczuk2020x,kurenkov2020visuomotor,novkovic2020object,danielczuk2019mechanical} or can be arranged following some underlying structure\cite{murali20206,sui2017goal,sui2020geofusion}. % repeated?
% , like how objects are arranged in a kitchen pantry where objects are stacked on beside and on top of each other creating structured clutter, inherently favouring some arrangements over others. In \cite{murali20206} the authors investigate grasping strategies to retrieve target objects from such structured clutter. \sehoon{maybe this can be merged to the scene generation section.}
While these approaches learn to search for a given target object, a more generalized version of this problem is to identify and locate \emph{all} objects present in the scene. Estimating the \emph{state} of such structured cluttered scenes has been explored by prior works~\cite{sui2017goal,sui2020geofusion} by using passive and active perception techniques.
Another approach to tackle the problem is to get the scene to a state where all objects are clearly visible~\cite{poon2019probabilistic}. However, this may not always be feasible since additional space might not be available in some scenarios. 
% \deleted{Their approach also relies on a priori knowledge of the number of objects in the scene. }
% \deleted{Similar to~\cite{poon2019probabilistic}, our objective is also to estimate the underlying true state of the scene. However, we take a reinforcement learning based approach to understand scenes by rearranging objects in a confined bin, that can generalize to scenes with arbitrary number of objects. }
\paragraph*{Scene Generation}
% \deleted{Synthetic scene generation is an important area of research that has applications in computer vision, robotics and computer graphics.} 
The goal of scene generation is to create synthetic counterparts of realistic scenes that share resemblance at a structural and semantic level. 
A natural choice of representation for scenes is a \emph{scene graph}, which typically represents objects as nodes and uses the edges to represent relational attributes between objects~\cite{krishna2017visual,liu2014creating, wang2019planit}. A body of work in scene generation~\cite{zhou2019scenegraphnet,devaranjan2020meta,qi2018human} focuses on creating 3D scenes by first creating scene graphs and then instantiating the represented scene using a renderer. 
% \deleted{Such a representation is used by \cite{zhou2019scenegraphnet} to determine which object can be added to an existing scene while satisfying the semantic narrative. This requires the construction of a graph by sequentially adding specific nodes and edges at specific locations on an intermediate graph.}
% , such that it fits into the semantic narrative of the room (for example a television is more likely to be in a living room than a bathroom). Other approaches construct entire scenes instead of just augmenting an already existing synthetic scene. This requires the construction of a graph by sequentially adding specific nodes and edges at specific locations on an intermediate graph. \sehoon{Maybe we can cut a few sentences, because it is not about robotics.}
Formal systems~\cite{chomsky2009syntactic} offer a promising approach to generate scene graphs. A formal system is defined by a set of non-terminal nodes, terminal nodes and production rules that define transformation from one sequence to another.
% \deleted{These systems were originally formulated to tackle natural language processing and linguistic problems, but can be modified to represent graphical structures.} 
Production rules can then be defined to add, remove or modify nodes and edges in the graph. Scene grammar was utilized by prior work 
% {In \cite{lau2011converting}, the authors define such a formal system to disintegrate furniture models and represent them using primitive shapes. More recently,}
~\cite{devaranjan2020meta,purkait2020sg} to generate simulated scenes that mimicked the structure and semantics of a real-world dataset. In contrast, we take a self-supervised approach to scene generation that is grounded in physical plausibility, by building scenes with cluttered but stable object arrangements in a \textit{physics simulator}.
%%SH: I deleted the below since it is a bit out of context
% Scene grammar has also been explored in the robotics research community to learn skeletal structures that can effectively navigate difficult terrain \cite{Zhao2020robogrammar}. 
% They use a recursive graph grammar that is optimized by a reinforcement learning agent to generate a skeleton that can traverse the terrain efficiently. In our problem we use a similar recursive grammar to define our scenes but we want our scene generator to create a distribution of adversarial scenes instead of just one scene containing hidden objects.
In the work of Sui~\etal~\cite{sui2017goal}, scene graphs are used along with a particle filter to represent multiple hypothesis of the underlying object-object relationships in the scene. They however assume that all the objects in the scene are visible and identifiable.  In our case, the scenes frequently contain heavily occluded objects that even a state-of-the-art object detector will not be able to identify. Instead, we take an RL-based approach and learn manipulation strategies that can efficiently uncover (dis-occlude) these objects.
% \vspace{-2mm}
\paragraph*{Object Rearrangement}
Object rearrangement in the context of robotic manipulation has been investigated by prior works~\cite{batra2020rearrangement,zeng2020transporter,danielczuk2020object}. In a typical object rearrangement scenario, the agent is given a goal state (either explicitly or implicitly, such as organizing a pile of grocery items in a pantry). Similarly, in our work we train an agent to rearrange with the objective of revealing hidden objects. However, in contrast to prior approaches in object search~\cite{danielczuk2020x,danielczuk2019mechanical}, we \emph{do not require an auxiliary bin} to discard occluding objects. This provides our agent two advantages: \emph{(i)} in real world environments, the robot may not have access to additional space where it can discard objects, giving our approach a competitive edge; and, \emph{(ii)} at the end of its sequence of interaction, our robot knows the location and identity of every object present in the scene and can retrieve multiple objects without any additional exploration. 
% This makes our work appropriate for real world tasks that might require adapting to continuously varying goals.
% While we do not have a singular goal state in our problem, we have distribution of states, which when reached give the robot a complete understanding of the scene.
% Instead, our goal is to completely understand the given scene, via object rearrangement with multiple interactions.
% While rearranging objects our robot executes controlled manipulations that moves the state of the scene through a series of partially observed states that gradually pull together the robot's belief about the world and the true underlying state.

% \vspace{-2mm}
\section{Graph-based Scene Generation and Exploration}
% !TEX root = main.tex
\begin{figure*}[t]
\vspace{2mm}
\centering
\includegraphics[width=0.9\linewidth]{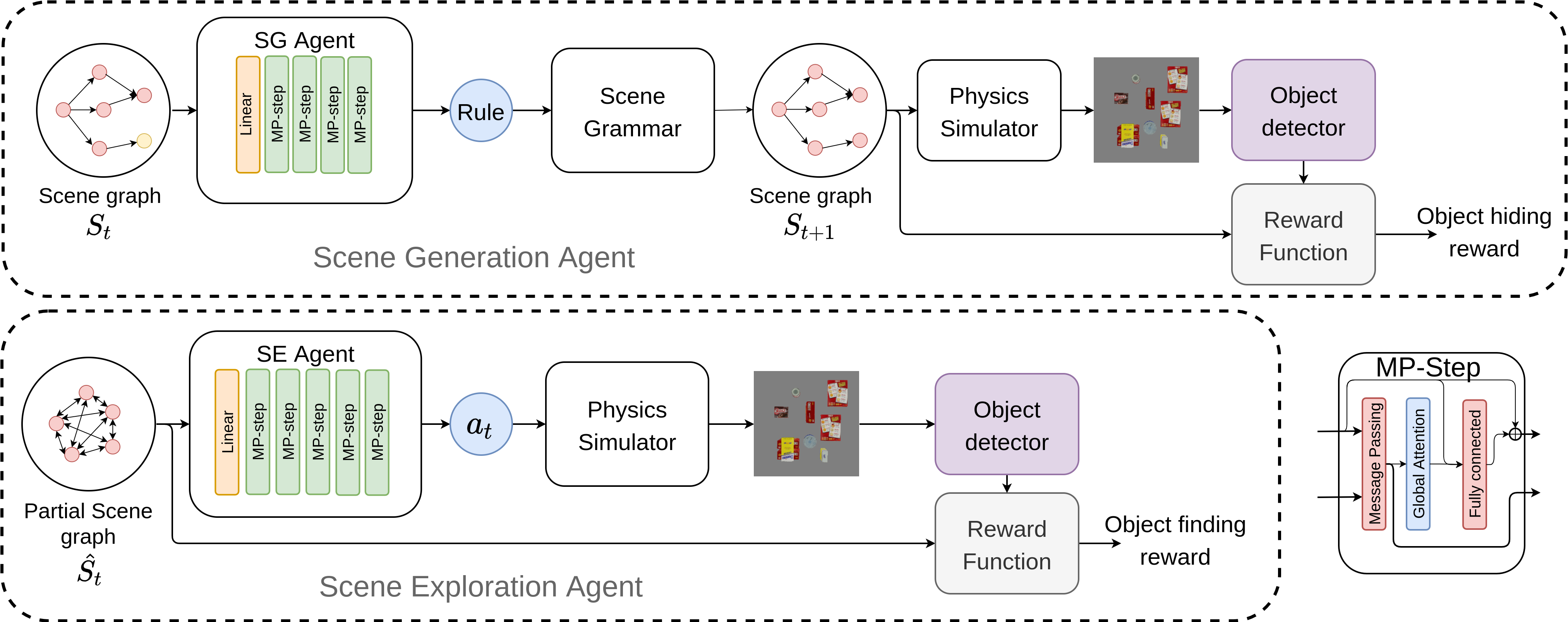}
\label{fig:architecture}
\caption{Overview of our method. We train two agents with conflicting objectives while ensuring that the scene remains stable. The Scene Generation agent (SG) tries to hide more objects in the scene while the scene exploration agent (SE) tries to find them. Each MP-step consists of a message passing layer followed by Global Attention~\cite{li2015gated} with a skip connection.
% \sehoon{delete MP-Step and discuss in the text.}\sehoon{The SE agent looks like getting the privileged information of the entire scene graph. Change the image to see the partial graph.}
\vspace{-4mm}
}
% \sehoon{This is good. Can you add captions to the graph (with circles). People will not know it without reading figures.}}
\end{figure*}
The core idea of our method is to utilize scene graphs as a unified representation that generate and rearrange cluttered scenes with many objects. 
Our method consists of two agents: \emph{(i)} a Scene Generation Agent that exploits a scene grammar and constructs scene graphs to represent difficult, physically stable and cluttered multi-object scenes; and, \emph{(ii)} a Scene Exploration Agent that starts with scenes generated by the Scene Generation Agent and outputs a sequence of efficient manipulation strategies that uncovers hidden objects and reveals the underlying scene graph. 
We represent both these agents as
% Message Passing Neural Networks 
MPNNs and train them with RL to maximize their respective objectives.
% scene graph is seamlessly used in adv and exp
% \vspace{-2mm}
\subsection{Scene Grammar}
In our approach, both the scene generation and scene exploration agents need to handle various scenes with multiple objects, thereby requiring the algorithm to efficiently represent and synthesize different configurations. To model the space of possible scenes an agent might encounter, we propose a \emph{scene grammar} defined by a set of production rules that represent transformations from one scene graph to another. These rules are sequentially applied to add or remove objects to a given scene graph, thereby letting us control the structure and composition of a scene.
% {To this end, we define a \emph{scene grammar} that can create \emph{scene graphs} by sequentially applying production rules that add and position objects in the right locations and orientations.} 
Our scene grammar is recursive and contains production rules that recursively add structures to the graph to mimic stacking objects on top of and/or next to each other. We define the scene grammar $\mathcal{G}$ as follows:
\begin{equation}
\label{eq:scene_grammar}
    \mathcal{G} = (N_{NT}, N_{T}, E, R, S)
\end{equation}
where $N_{NT}$ are the non-terminal nodes, $N_{T}$ correspond to the terminal nodes, and $S$ is the start graph. 
$E$ represents the set of edges available in the grammar. 
Terminal attributes represent \emph{real} aspects of the scene such as object type or orientation, while the non-terminal attributes are intermediate entities that help with the construction of the graph. 
For a graph to be simulated as a scene, all its attributes must be terminal. $R$ represents the production rules that are applied to transform the graph. 

\begin{figure*}[t]
\vspace{2mm}
\centering
\includegraphics[width=0.99\linewidth]{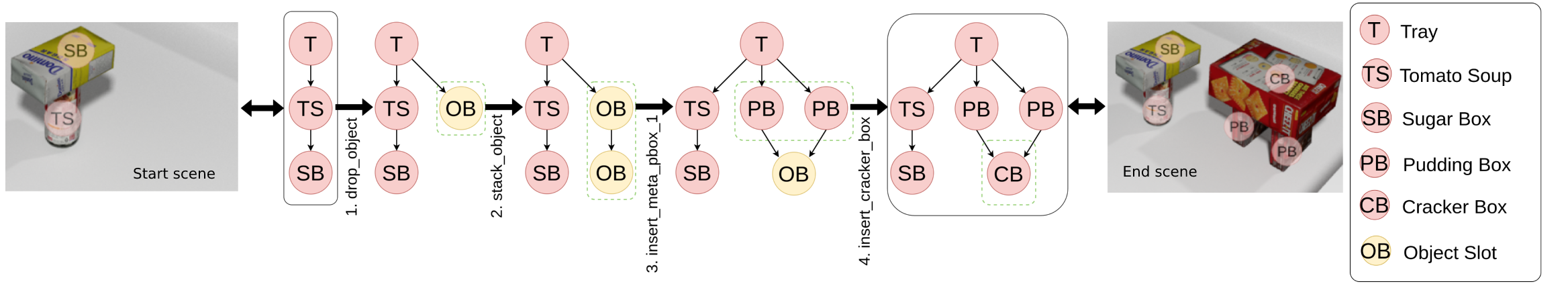}
\caption{We show an example to demonstrate our proposed scene grammar by applying a sequence of rules to a scene graph. The agent starts with a scene that contains a sugar box stacked on top of a tomato soup can on a tray. The agent then applies a series of rules to add more objects, namely two pudding boxes and a cracker box, to the scene.
}
% \caption{We show an example to demonstrate our proposed scene grammar by applying a sequence of rules to a scene graph(shown in a topologically sorted representation). The object and non-terminal nodes are represented by red and yellow nodes respectively. The gray nodes are terminal nodes that do not represent real objects but exist to simplify the grammar. The dashed lines and solid lines represent non-terminal and terminal edges respectively. The agent starts with a scene that contains a tomato soup can (TS) stacked on top of a sugar box (SB) on top of a tray (T). (E) denotes an \textit{end\_node} that prevents further stacking along the path of the graph. This is a terminal graph and can be rendered with the simulator as shown in the image in the top right. The agent first applies the \textit{drop\_object} rule. This inserts an \textit{object\_slot} (OB) and an \textit{open\_slot} (OP) into the scene graph. Next we use \textit{stack\_object} to stack another \textit{object\_slot} (OB) on top of the one we inserted in the previous step. We use \textit{insert\_meta\_pudding\_box\_1} that inserts two pudding boxes (PB) next to each other on the tray in a predefined orientation. The \textit{dummy\_node} (D) is used to simplify the grammar by abstracting \textit{meta} objects with a single node, on top of which other objects can be stacked. Next we insert a \textit{cracker\_box}(CB) on top of the two pudding boxes and close the path so that no further objects can be added on top.}
\vspace{-4mm}
\label{fig:scene_grammar}
\end{figure*}

Each production rule is represented with two graphs, $G_L$ and $G_R$, that define the transformation it can perform. 
To apply a rule, we first search (using BFS) for a subgraph  in the input graph that is isomorphic to $G_L$ and then replace it with $G_R$.
% \begin{equation}
%     G_L \rightarrow G_R
% \end{equation}
Both $G_L$ and $G_R$ can contain terminal and/or non-terminal attributes. 
% This allows us to define production rules that can convert a sub-graph $G_L$ with terminal/non-terminal nodes in to $G_R$ containing non-terminal nodes, letting us recursively keep adding any arbitrary number of additional nodes.
\added{In our grammar we use a non-terminal node called \textit{object\_slot} as a placeholder that can later be substituted by object nodes. This design choice disentangles the structure of the graph from the actual objects it will be instantiated with, leading to a reduction in the number of grammar rules required.

Consider the example in Figure~\ref{fig:scene_grammar}. We start out with a scene graph that contains two objects, a tomato soup can and a sugar box. To add new objects near the existing stack, we first apply the \textit{drop\_object} rule that creates a new \textit{object\_slot} attached to the tray. Next, we apply \textit{stack\_object} to stack another \textit{object\_slot} on-top of this node. 
Note that the resulting scene graph has non-terminal nodes (\ie two \textit{object\_slot} nodes), and cannot be simulated as a scene. 
We then apply rules that substitute the \textit{object\_slot's} with actual objects that can be simulated in a scene. First we apply an \textit{insert\_meta\_pbox\_1} rule that inserts two pudding boxes, right next to each other, upon which more objects can be stacked. Finally, we apply an \textit{insert\_cracker\_box} rule that substitutes the topmost \textit{object\_slot} with a cracker box. This results in a scene graph that just contains terminal nodes that can be simulated as shown.}
% In our grammar definition, we have two non-terminal nodes: \textit{object\_slot} which can be substituted by any object available to the agent and \textit{open\_slot} which can be substituted by either an \textit{object\_slot} or an \textit{end} node. 
% The \textit{end} node is a terminal node that stops any further node modification to that region of the graph. When stacking multiple objects next to each other, we use a \textit{dummy\_node} that acts as a single-node end point on which more objects can be stacked (as shown in Fig. \ref{fig:scene_grammar}). 
% The remaining terminal nodes are all object nodes that correspond to real world objects. Our grammar also includes a non-terminal edge called \textit{primitive} that can be substituted with any of the terminal edges that represent the actual orientation of objects in the world. See Figure \ref{fig:scene_grammar} for an illustrative example. 
\added{For more details about implementation, see our project page} \footnote{\url{https://www.kniranjankumar.com/projects/5_clutr/}}. %\lfoot{\href{link}{https://www.kniranjankumar.com/projects/5_clutter_exploration/}.}
% \sehoon{place the link to the project page on your website, not Arxiv}
% \sehoon{Can we revise the rule and majorly rewrite this section? Basically, the rule is very complicated. I would not introduce the concept of End node. All the leaf nodes can be considered as "E" nodes. "D" is not the terminal node. Merge the concept of Object slot and Open spot. We don't need to distinguish two edges. Even not sure S is defined here. I know this is very different from the actual implementation, but people would prefer simpler notations. The core idea will be the same. }
% \sehoon{Try to explain with concrete examples as much as possible.}\niranjan{How about this?}
% \sehoon{I will revisit this section when you add a figure.}

% \vspace{-2mm}

% The goal of the adversarial agent is create a distribution of difficult and stable scenes that resemble real world structured clutter. 
\subsection{Scene Generation Agent}
In the previous section, we defined a scene grammar that can represent arrangements of objects in a scene. But not all scene graphs that can be generated, would actually be stable in the real world. In addition, we want to generate scenes that have many hidden objects that escape detection from a vision-based perception system. 
This raises the question: \emph{how can we procedurally generate difficult, stable and cluttered scenes from the set of production rules?} 

We model the process of generating a scene graph using our scene grammar as a Markov Decision Process (MDP) defined by the 5-tuple; $ \mathcal{M} : \langle \mathcal{S},\mathcal{A},\mathcal{T}, \mathcal{R},\gamma \rangle$ where $\mathcal{S}$, $\mathcal{A}$ and $\mathcal{R}$ are the state space, action space and reward function respectively. $\mathcal{T}$ is a transition function that determines the next state given the current state and action. $\gamma \in (0,1)$ is the discount factor. We define how these terms relate to the scene generation procedure below.

\textbf{States}: We define the state at time-step $t$ to be the corresponding scene graph at $t$. This graph can contain both terminal and non-terminal nodes. Each node feature is defined to be a combination of the class the object belongs to, a boolean flag that denotes whether the node has been simulated and rendered in the environment and the position of the object, if it has been rendered in the scene. The edges in this graph represent the orientation of the child node.

% \textbf{Actions}: We define the action space to be a the set of all production rules $R$ defined in our scene grammar (Equation~\ref{eq:scene_grammar}).
\textbf{Actions}: We define the action space to be the set of all production rules $R$ (we use $30$ rules in our implementation). %{see Figure \ref{fig:all_production_rules}}).

\textbf{Reward}: The reward function for our agent consists of two components. A stability penalty $r_s$ that penalizes the agent if the constructed scene is physically unstable. An uncertainty reward $r_u$ that is proportional to the number of objects the agent is able to hide in the scene.
\begin{equation}
\label{eq:reward}
    \mathcal{R}(s) = w_s r_s + w_u r_u
\end{equation}
We use constant weights $w_s = -1$ and $w_u = 0.5$ for all our experiments.
The transition function $\mathcal{T}$ is determined by possible transitions induced by the scene grammar. 

%\sehoon{Summarize this paragraph too. Simply mention that we need a distribution, not a single instance.}\niranjan{fixed}
% The goal of the Scene Generation Agent is to navigate the state space, identify high-reward regions and learn a policy that can apply a sequence of actions sampled from $R$ to get from $S$ (start graph) to those regions. 
In addition to maximizing the above objective, we also want to generate high entropy policies to ensure that that we create a \textit{distribution} of cluttered scenes instead of just one that the Scene Exploration Agent can easily over-fit to. To this end, we exploit a maximum entropy RL based approach to this problem. We learn a policy with a MPNN architecture using Soft-Q learning~\cite{haarnoja2017reinforcement} to output a categorical probability distribution over production rules available at a given state.
\subsection{Scene Exploration Agent}\label{ssec:SEagent}
The next component of our approach involves training an agent to discover hidden objects and estimate the underlying state of the scene. We formulate this problem as a Partially Observed Markov Decision Process (PoMDP) where the agent does not know the true underlying state of the scene $\vc{s}$ and only has access to a partial observation $\vc{o}$. 

Consider a scene with $N$ objects (unknown during test time) represented by the set $\mathbf{S}:\{0,1, \dots N\}$, where $0$ represents the ground and $\{1, \dots N\}$ represents objects present in the tray. Since the scene can have occluded objects, we define a set of currently visible objects $\mathbf{V}_t \subseteq \mathbf{S}$, which contains all the objects detectable by the agent at time $t$, and a set of historically identified objects $\mathbf{K}_{t} \subseteq \mathbf{S}$, which contains all the objects the agent has seen \textit{at least} once. At time $t$, the agent updates $\mathbf{K_t}$ using $\mathbf{K_t} = \mathbf{K_{t-1} \cup \mathbf{V}_t}$.
% \begin{equation}
%     \mathbf{K_t} = \mathbf{K_{t-1} \cup \mathbf{V}_t}
% \end{equation}
% \sehoon{This equation can be merged to the text.}
The agent interacts with the scene until $\mathbf{K}_t \cup \{0\} = \mathbf{S}$. Given this premise, we define a PoMDP for the structured clutter search problem with the 7-tuple $\langle \mathcal{S},\mathcal{A},\mathcal{T}, \mathcal{O}, \Omega, \mathcal{R},\gamma \rangle$: 

\textbf{States} ($\mathcal{S}$): The underlying state of the scene containing $N$ objects, at a given timestep $t$, is a scene graph $S_t$ with objects represented as nodes and object-object relationships represented as edges. The node for each object $i \in \left[1,N\right]$ contains the following: Pose $p_i$, the class ID, a boolean flag that denotes if the object has ever been detected by the robot. 

\textbf{Actions} ($\mathcal{A}$): A tuple $(x,y)\mid x \in \mathbf{V_t}, y \in \mathbf{V_t} \cup \{0\} $ where object $x$ represents the object that is to be picked up and $y$ represents the object on top of which $x$ will be placed.

% \textbf{Observations} ($\Omega$): \replaced{Image}{An image obtained} from overhead RGBD camera.
\textbf{Observations} ($\Omega$): A partial scene graph $\hat{S_t}$ that contains objects seen by the agent, \ie objects in $\mathbf{K}_t$. Since, object-object relationships are unknown during test time, we define $\hat{S_t}$ as a fully connected graph with pose $p_i$ and class ID as the node features.

%\sehoon{This is a major mistake from my misunderstanding. I guess our observation is not image, the positions of the visible object. How do you represent them? Another scene graph? Define the partial scene graph precisely.}\niranjan{fixed}
%, poses of all objects in $\mathbf{V}_t$ and the partial scene graph $\hat{S}$ with nodes $n \in \mathbf{V_t}$ and edges $e:(i,j) \mid i,j \in \mathbf{V_t}$.
\textbf{Reward} ($\mathcal{R}$): The agent gets a reward of $r_{d}$ for every newly detected object and a penalty of $r_s$ if the scene becomes unstable, same as Equation \ref{eq:reward}. \sehoon{let's use the word penalty for $r_s$. Please update all accordingly.} \niranjan{fixed}

$\mathcal{T}$, $\mathcal{O}$ denote the transition function $p\left(\vc{s}_{t+1} \vert \vc{s}_t,\vc{a}_t \right)$ and the observation model $p\left(\vc{o}_t \vert \vc{s}_t \right)$ respectively, with $\gamma$ being the discount factor. With this setup, we train an agent to discover all the objects present in the scene by rearrangement. Such an agent should select actions that maximize the expected discounted sum of rewards defined by:
\begin{equation}
\label{eq:cumulative_reward}
J=\mathbb{E}_{p(\tau)}\left[\sum_{t=1}^{T} \gamma^{t-1} r_{t}\right]
\end{equation}
where $p(\tau)$ is a distribution over trajectories $\tau$ 
% \deleted{:($\vc{s}_{0},\vc{a}_{0},\vc{s}_{1},\vc{a}_{1}, \dots$) }
imposed by the policy. However, for a PoMDP we only have access to $\vc{o_t}$ and not the underlying state $\vc{s_t}$ during test time.

\textbf{Training with Privileged Learning:}
In our experience, it is not straightforward to solve the given PoMDP using a na\"ive end-to-end learning approach.
Inspired by the work of \cite{chen2020learning}, we propose a solution that uses a privileged agent to first solve the problem and then distill the network into a student agent that can mimic the actions taken by the former.

%\sehoon{It will be great if we can make the below more concise. (1) Define the teacher observation and contrast against the student observation. (2) Explain that the teacher can be solved with deep RL. (3) Then explain the behavior cloning by defining the loss function. That is it. Don't say too much details. Do you use a DAGGER algorithm?}\niranjan{How about now? No I just use vanilla BC}
% \vspace{-2mm}
Consider an alternate problem: \emph{how should an agent with complete knowledge of the scene rearrange objects such that every object gets revealed during the episode?}
The agent in this case gets access to the complete scene graph $S_t$ (instead of the partial graph $\hat{S_t}$) as input. Since it has access to the underlying state, we can formulate our problem as an MDP and solve it using traditional RL methods. We use a MPNN architecture to model our policy and train it with an actor-critic framework to maximize Equation \ref{eq:cumulative_reward}.
% This can be formulated as a well defined MDP where the agent has to step through a sequence of states by rearranging the objects in the scene graph, so that every object gets revealed at least once.
% To solve this MDP, we train a policy to output the optimal action $\vc{a_t}$ conditioned on the true underlying state of the scene $S_t$.  
% Since this policy gets the complete scene graph as input at every step, it can decide how to rearrange nodes in the graph so that every object becomes visible at least once during exploration.
We call this agent the privileged Scene Exploration agent $\pi^{*}(\vc{a}\vert \vc{s})$. We then collect a dataset $\mathcal{D}$ consisting of tuples $\langle \hat{S}, \vc{a}^* \rangle$ by running the privileged policy. Subsequently, we use Behaviour Cloning to train the student agent $\pi(\vc{a}\vert \vc{s})$ by minimizing the following objective.
\begin{equation}
\label{eq:BC_loss}
L =  -\sum_{\vc{a}^{*},\hat{S_t} \in \mathcal{D}} \log \pi(\vc{a}^{*}\vert \hat{S_t})
\end{equation}
% \niranjan{Update this}
where $a^{*}$ is the action taken by $\pi^{*}$ for state $S_t$. 

% . Once we train this agent we can look back to our original problem of solving the PoMDP.
% We collect a dataset consisting of 3-tuples $\langle \vc{s_t}, \hat{S}, \vc{a} \rangle$ by running the privileged policy $\pi^{*}(a\vert s)$. Subsequently, we train a student policy $\hat{\pi}(a\vert s)$ that takes as input the partial scene graph $\hat{S}$ extracted from the RGBD image and outputs the action predicted by $\pi^{*}(a\vert s)$. 
% $\hat{S}$ is a fully connected graph that consists of the objects that can be detected by the object detector ($\mathbf{K_t}$) and their respective poses.
% We train this student policy with behaviour cloning, maximizing the log-likelihood of the predicted action that matches the ground truth action taken by the privileged agent. 
This forces the student policy to discover patterns in the arrangement of \emph{visible} objects, namely priors about physical plausibility, giving clues about where to find hidden objects. Once trained, the student policy operates with incomplete information about the scene and takes actions that have a high likelihood of uncovering hidden objects in the scene.

% \vspace{-2mm}
\section{Experiments}
% {-2mm}
% \examplesimage
\begin{figure}[t]
\vspace{2mm}
\centering
\begin{subfigure}{0.47\linewidth}
\includegraphics[width=0.95\linewidth]{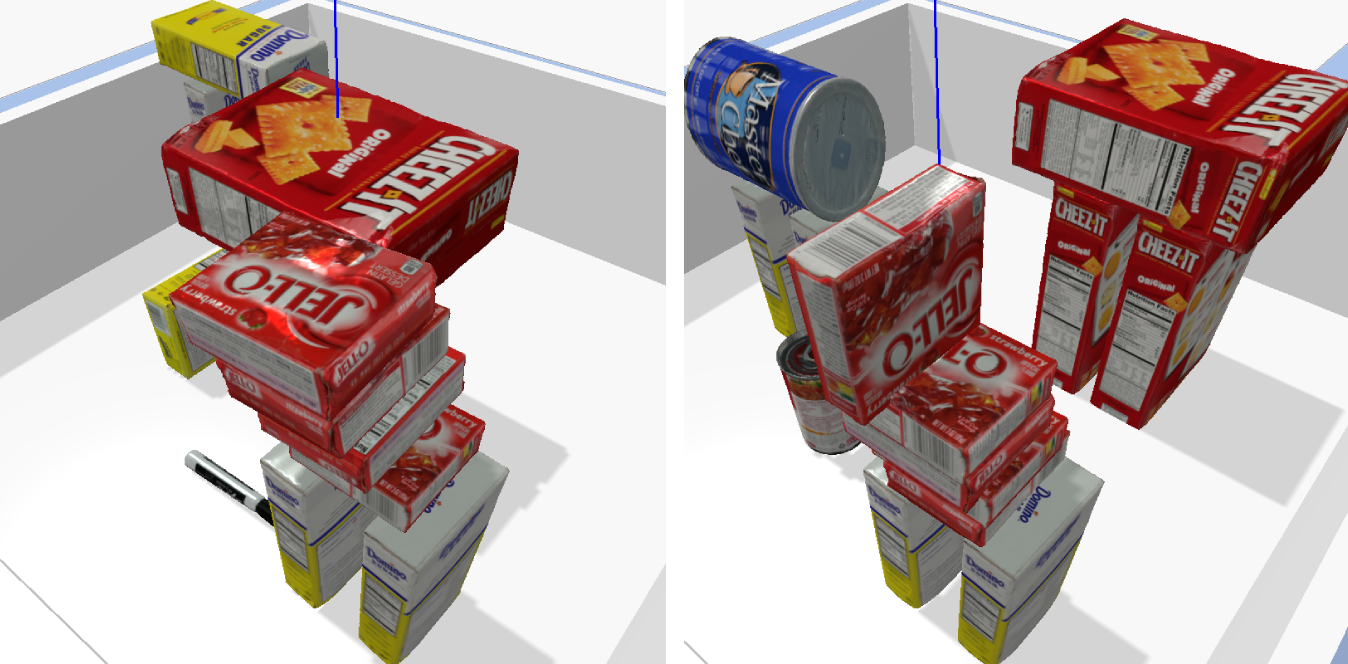}
\caption{}
\label{fig:scene_generation_7obj}
\end{subfigure}%
\begin{subfigure}{0.47\linewidth}
\includegraphics[width=0.95\linewidth]{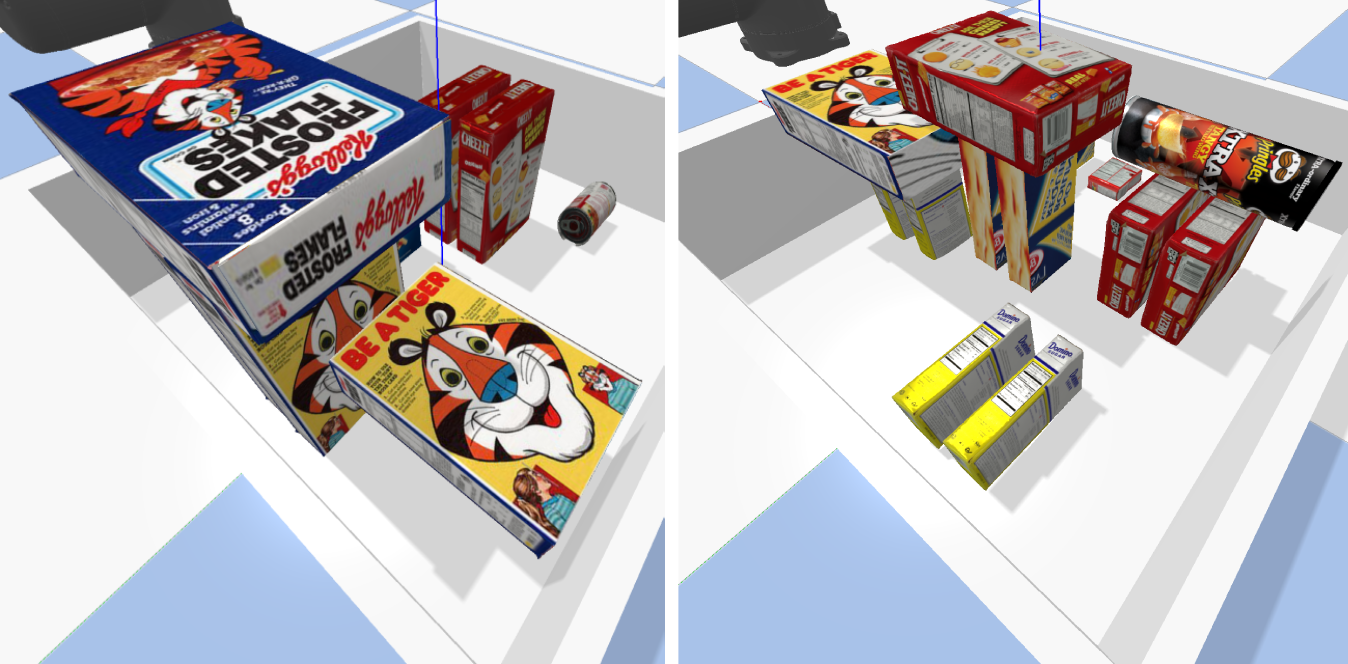}
\caption{}
\label{fig:scene_generation_14obj}
\end{subfigure}%

\caption{Examples of scenes generated by the (a) Scene Generation Agent trained on $7$ objects (b) Scene generation Agent trained on $14$ objects.
\sehoon{(b) does not seem particularly more complicated than (a). Can you select the most complicated scene, hopefully with 25 objects?}
}
\label{fig:scene_generation}
\vspace{-4mm}
\end{figure}
We evaluate our method in simulation using iGibson~\cite{shen2020igibson} and Pybullet~\cite{coumans2019pybullet} and on real world clutter using a UR10 robot. We utilize objects from the YCB \cite{calli2015ycb} dataset to create multi-object structured clutter that an interactive agent can explore. 
% We discuss the scene graph representations that we use as input, the model architecture of our agents and details about the object detector in the appendix.
We train Scaled-YOLOv4~\cite{wang2020scaled} network and use it as the object detector. To train the object detector, we first create a training dataset of simulated scenes by dropping $20$ objects (randomly sampled from the set of objects defined previously) into a bin and generate 100k annotated images. 
We then augment this dataset with 10k images of scenes generated with the scene grammar. 
In each case, we capture an image of the simulated scene from the overhead RGBD camera and record the bounding box annotations for all objects in the scene. 
Our simulated scenes consist of seven objects from the YCB dataset:
% {We evaluate our agents in a simulated iGibson world with seven types of objects from the YCB dataset:}
Cracker box, Pudding box, Master chef can, Tomato soup can, Sugar box, Gelatin box and Large marker. 
These objects vary both in shape and size, presenting abundant opportunities for objects to be hidden efficiently. 
We further demonstrate the scalability of our method, by running experiments with an extended version of this set that has $7$ additional kitchen items ($2 \times$cereal, granola, pasta, lasagna, chips boxes and a pringles can).
The Scene Generation Agent strategically creates stable scenes with cluttered arrangements of these objects, which the Scene Exploration Agent explores through rearrangement.
% We use a simulated UR10 robot fitted with a suction gripper to move objects around in the bin as instructed by our exploration policy.
% Move network architecture to appendix?
% \begin{figure}[t]
% \centering
% \includegraphics[width=0.95\linewidth]{images/example_scene_generation_v2.png}
% \caption{Examples of scenes generated by running our Scene Generation Agent.}
% \label{fig:example_scenes}
% \end{figure}
%  \vspace{-2mm}
\begin{table}
\vspace{2mm}
\centering
\begin{tabular}{|c|c|c|c|c|c|c|}
\hline
\multirow{3}{*}{Num. Nodes} & \multicolumn{3}{c|}{Num. Objects Hidden} & %
    \multicolumn{3}{c|}{Stability Rate}\\
\cline{2-7}
%  & \multicolumn{2}{c|}{Value} & \multicolumn{2}{c|}{Value} & \\
% \cline{2-5}
 & RG & SG & SG$^*$ & RG & SG & SG$^*$\\
\hline
% 10 & 0.94 & 2.09 & \textbf{4.47} & 0.84 & \textbf{0.99} & 0.97 \\
% 15 & 1.04 & 2.38 & \textbf{4.99} & 0.64 & \textbf{0.99} & 0.85\\
% 20 & 1.19 & 2.28 & \textbf{6.46} & 0.55 & 0.93 & \textbf{0.97}\\
% 25 & 1.02 & 2.43 & \textbf{6.55} & 0.41 & 0.81 & \textbf{0.88}\\
10 & 1.13 & 2.12 & \textbf{4.62} & 0.84 & \textbf{0.99} & 0.97 \\
15 & 1.64 & 2.41 & \textbf{5.88} & 0.64 & \textbf{0.99} & 0.85\\
20 & 2.18 & 2.46 & \textbf{6.67} & 0.55 & 0.93 & \textbf{0.97}\\
25 & 2.51 & 3.01 & \textbf{7.45} & 0.41 & 0.81 & \textbf{0.88}\\
\hline
% \label{tab:scene_generator_results}
\end{tabular}
\caption{\label{tab:scene_generator_results} Performance of the Scene Generation Agent (SG) with graph size $=15$ nodes and Scene Generation Agent$^*$ (SG$^*$) with graph size $=25$ nodes, compared to Random Generation Agent (RG). Stability rate is defined as the fraction of simulated scenes that are physically stable.}
 \vspace{-2mm}
\end{table}

\begin{table}
\vspace{2mm}
\centering

\begin{tabular}{|c|c|c|c|c|c|c|c|c|} 
\hline
Num.~  & \multicolumn{4}{c|}{Success Rate}            & \multicolumn{4}{c|}{Num objects found}       \\ 
\cline{2-9}
Nodes  & RE    & LF   & SE$^{r}$      & SE$^*$    & RE   & LF   & SE$^{r}$      & SE$^*$      \\ 
\hline
10     & 0.05  & 0.88 & \textbf{0.92} & 0.91          & 0.76 & 2.93 & 3.06          & \textbf{3.13}  \\
15     & 0.03  & 0.58 & \textbf{0.84} & 0.82          & 1.15 & 3.33 & \textbf{4.40} & 4.27           \\
20     & 0.00 & 0.32 & 0.36          & \textbf{0.64} & 0.93 & 3.64 & 4.24          & \textbf{4.71}  \\
25$^*$ & 0.00  & 0.10 & 0.52          & \textbf{0.91} & 0.90 & 2.70 & 4.87          & \textbf{5.82}  \\
\hline
\end{tabular}
\caption{Performance of Scene Exploration Agent (SE$^*$) trained with scenes generated by SG*, compared to the Random Exploration Agent (RE), Largest First Agent (LF), and an agent (SE$^{r}$) trained with scenes generated by RG. SE$^*$ shows higher success rates, particularly for larger scenes.}
\label{tab:scene_explorer_results}
% \vspace{-4mm}
\end{table}

\begin{table}[!ht]
\centering
\begin{tabular}{|c|c|} 
\hline
Object set                                                          & Objects hidden/found  \\ 
\hline
$7$ objects                                                              & $7.45/5.82$             \\ 
\hline
$14$ objects                                                            & $6.03/5.33$             \\ 
\hline
% \multicolumn{1}{l}{} & \multicolumn{1}{l}{\begin{tabular}[c]{@{}l@{}}\\\end{tabular}} & \multicolumn{1}{l}{}  \\
% \multicolumn{1}{l}{} & \multicolumn{1}{l}{\begin{tabular}[c]{@{}l@{}}\\\end{tabular}} & \multicolumn{1}{l}{} 
\end{tabular}
\caption{Scalability analysis: we test our approach on two object sets, containing $7$ and $14$ objects. Note that in each case, we train the Scene Generation agent to generate scene graphs with $25$ nodes using the available object types.}
\label{tab:scalability}
\vspace{-4mm}
\end{table}
\begin{figure*}[t]
\vspace{2mm}
\centering
\begin{subfigure}{0.66\linewidth}
\includegraphics[width=0.95\linewidth]{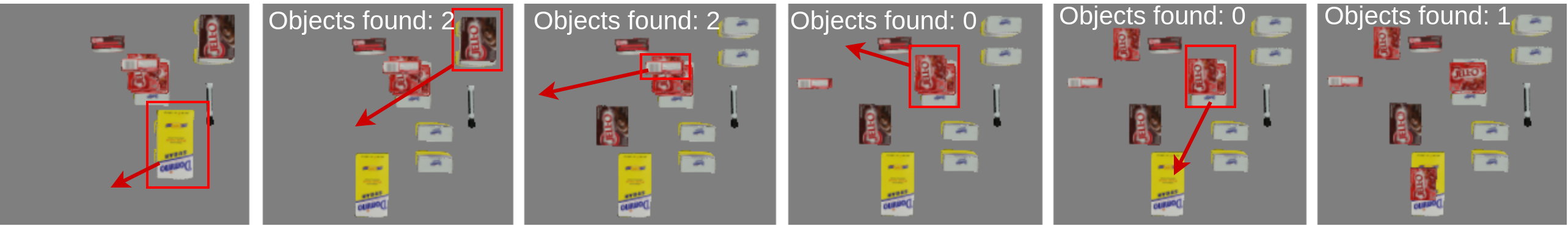}
\caption{}
\label{fig:simulated_exploration}
\end{subfigure}%
\begin{subfigure}{0.33\linewidth}
\includegraphics[width=0.95\linewidth]{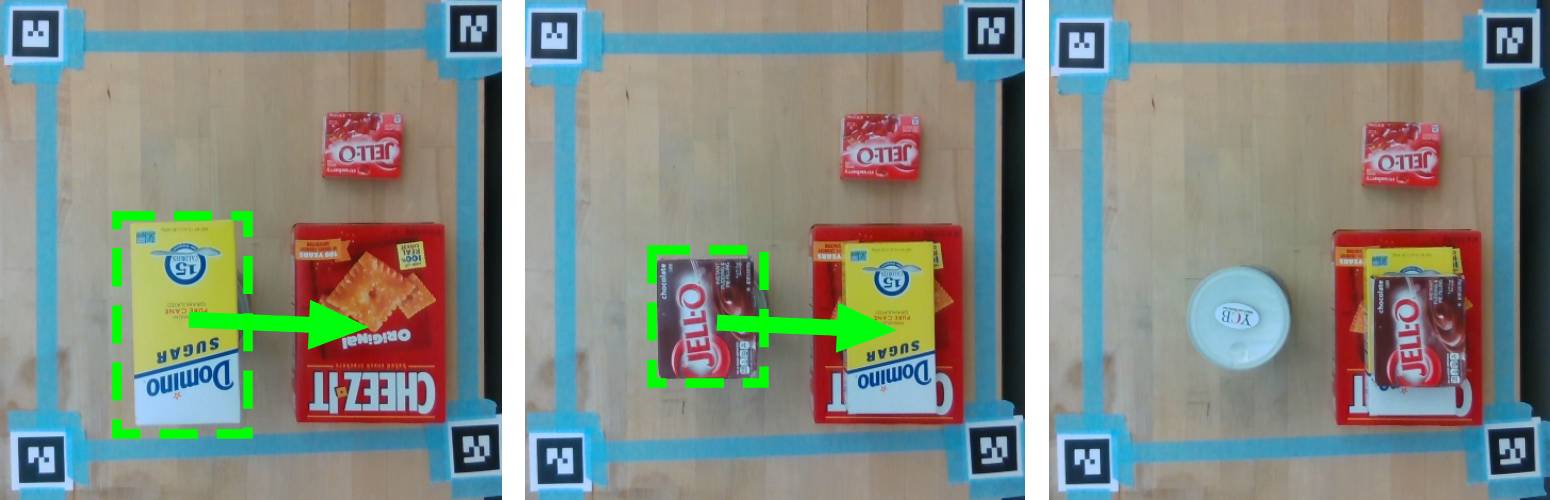}
\caption{}
\label{fig:real_exploration}
\end{subfigure}%

\caption{Example actions performed by the Scene Exploration Agent (a) In simulation (b) Real world. In each case, the agent rearranges the scene and discovers hidden objects ($5$ in the simulated scene and $2$ in the real world clutter.)
% \sehoon{If we have more spaces, we can add more images!}
}
\label{fig:scene_exploration}
\vspace{-3mm}
\end{figure*}
% \deleted{We want our agent to create stable cluttered scenes while maximizing the number of hidden objects. }
%\sehoon{Don't talk about the implementation details. Move to the method section if necessary.}\niranjan{fixed}

\textbf{Scene Generation agent:} The Scene Generation Agent takes a scene graph as input and outputs a distribution over the set of production rules $R$. 
% We represent nodes of this scene graph as the concatenation of a one-hot vector encoding the node type, a boolean flag \textit{is\_simulated} that represents if the node has been simulated, and the $3$ dimensional position of the node if \textit{is\_simulated} is set to be True. The edges are represented as one-hot vectors that specify the type of the edge. 
Before feeding the scene graphs into a network, we first embed each node using a fully connected (FC) layer followed by a non-linearity (Leaky-ReLU), resulting in a vector of  $32$ dimensions. We use a similar network to embed the edges into a vector of size $3$.  
%
%\sehoon{Is this different from MPNN?}\niranjan{Yes, a little bit. We have global attention and skip connections}
The architecture of our policy is similar to the one used by~\cite{janisch2020symbolic} for BlockWorld, but with four message passing layers. We pass the output of the final layer through a FC layer to generate Q-values for all the rules in our graph grammar. We train the agent for a total of $640k$ steps starting with a learning rate of $10^{-3}$ and halving it every $128k$ steps.

We compare the performance of our trained Scene Generation (SG) agent to a Random Scene Generation policy (RG) that samples uniformly from the set of feasible production rules. To evaluate the performance of our scene generators, we define two metrics: \emph{(i)} the number of objects they successfully hide; and, \emph{(ii)} the stability rate, which indicates the probability of generating physically stable scenes. 
We train two agents SG (trained to generate graphs with 15 nodes) and SG${^*}$ (trained to generate graphs with 25 nodes) and compare their performance to RG (Table \ref{tab:scene_generator_results}). 

We find that both our agents outperform the baseline random agent for the graph sizes they were trained on. 
SG$^*$ can generate scene graphs with $25$ nodes that contain an average of $7.45$ objects ($\approx5$ more than RE), with a stability rate of $88\%$ ($43\%$ more than RE). 
Interestingly, we also observe that our agents can generalize to graph sizes they were not trained to generate. 
For example, even though SG has never been trained to generate graphs with $10, 20$ and $25$ nodes, during test time, we can adjust the graph size to reliably generate stable cluttered scenes of different sizes. 
To test the scalability of our approach we train another Scene Generation agent on the extended version of the object set (containing $14$ objects) as described earlier.
% This new setup contains the $7$ objects from the YCB dataset plus an additional $7$ custom pantry objects as described earlier.
% (See project page \footnote{\url{https://www.kniranjankumar.com/projects/5_clutter_exploration/}} for details). 
We augment our grammar with an additional set of $19$ rules ($49$ in total) to support these additional objects and structures. This agent hides $6.03$ (stability rate $=70\%$) objects per scene when compared to a random agent, that only hides an average of $2.44$ (stability rate $9\%$) objects. 
We observe that in spite of doubling the object set size, the performance of our method is not degraded (Table~\ref{tab:scalability}).
A few examples of generated scenes are shown in Figure~\ref{fig:scene_generation}.

\textbf{Scene Exploration agent:} 
%\sehoon{Don't talk about the implementation details here. Move to the method section if necessary.} 
% \added{The input to the Privileged Scene Exploration Agent is also a scene graph, but with some modifications. We use replace \textit{is\_simulated} with an \textit{is\_seen} flag to indicate whether an object has been seen or not. The node features also contains the orientation of the object represented as quaternions. The edges of this graph are represented with a single boolean flag to denote if the two objects are connected. For the student Scene Exploration Agent we use a slightly different input representation, where the input is a fully connected graph of all the object nodes seen until the given point (as described in Section \ref{ssec:SEagent}).} 
We use architecture identical to \cite{janisch2020symbolic} for both Privileged and Student Exploration agents. 
% We train the Privileged Scene Exploration Agent for $150k$ steps using an actor-critic framework.
To evaluate the performance, we define the following metrics: \emph{(i)} success rate (SR), which denotes the fraction of scenes where all objects were found; and, \emph{(ii)} the average number of objects found per scene (OF). We first generate a dataset of $3k$ scene graphs each containing $25$ nodes using our Scene Generation agent (SG$^{*}$). We randomly sample a scene graph from this dataset to initialize the scene, but randomize the object locations on the tray. 

We compare our Scene Exploration Agent (SE$^*$) to three baselines: \emph{(i)} a Random Exploration Agent (RE), that randomly rearranges objects; \emph{(ii)} the Largest First agent (LF) that moves the largest objects in the scene to look under them, as described in Danielczuk~\etal~\cite{danielczuk2020x}; and, \emph{(iii)} the Scene Exploration Agent (SE$^{r}$) which is trained similar to SE, but on stable scenes generated by the random generator (RG) instead of the learned SG. We observe that the baselines could work for simpler scenes, but as complexity increases, our method provides a significant margin of improvement. Largest First agent (LF) performs well for graphs with $10$ or $15$ nodes, but fails on larger scenes.
While SE$^{r}$ and SE$^*$ perform comparably for small scenes, the importance of the Scene Generation Agent becomes evident with larger scenes. SE$^*$ has a Success Rate of $91\%$ which is a $75\%$ relative improvement over SE$^{r}$(SR$=52\%$). 
To test the scalability of our method, we train another Scene Exploration agent on scenes generated by the $14$ objects version of the Scene Generation agent. This agent is able to find all hidden objects in $64\%$ of the scenes outperforming LF ($56\%$) and RE ($3\%$). As shown in Table~\ref{tab:scalability}, our Exploration agent is able to reliably find objects despite doubling the object set size. An example exploration sequence is shown in Figure~\ref{fig:simulated_exploration}.
% \sehoon{Looooooong paragraph again. Split the paragraph, and try to start each paragraph with the key message.}
% \sehoon{Let's match the terms: SG-, SG, SE-, SE. Or SG, SG*, SE, SE*}\niranjan{Hmm. SG and SE are quite different though. SE is not trained on SG but on RG}
% ($\approx 1$ additional object)
% \deleted{We give the agent a maximum of $10$ steps to find all the objects. If the agent takes an action that causes the scene to destabilize during exploration, we terminate the episode early.}
% \deleted{As seen in }\ref{tab:scene_explorer_results}\deleted{, our trained Scene Exploration Agent is able to discover an average of $2.31$ objects per scene. 
% Further, it is able to find all hidden objects in $85\%$ of the scenes.}
% \deleted{In comparison, an agent that randomly rearranges the scene (RA) can only find an average of $0.79$ objects, with all objects being found in only $18\%$ of the cluttered scenes. Next, we train a Scene Exploration Agent on scenes generated by SG$^*$, that have more hidden objects per scene on an average. This agent is able to solve $72\%$ of the scenes and finds an average of $5.8$ objects in every scene, while a random agent can just find $0.92$ objects and does not solve any scene completely.}

% \newpage
% \realworldclutter

% \sehoon{We can mainly discuss SG*.}\niranjan{fixed}
% \sehoon{cannot find Figure 4. But consider splitting them, in the sense that the message for each subfigure is very different. (and maybe save the image.}\niranjan{fixed}
% \sehoon{Figure 4a might look better with quaterview. Not sure.}\niranjan{fixed}

\begin{figure}[t]
    \centering
    % \vspace{-10pt}
    \includegraphics[width=0.4\textwidth]{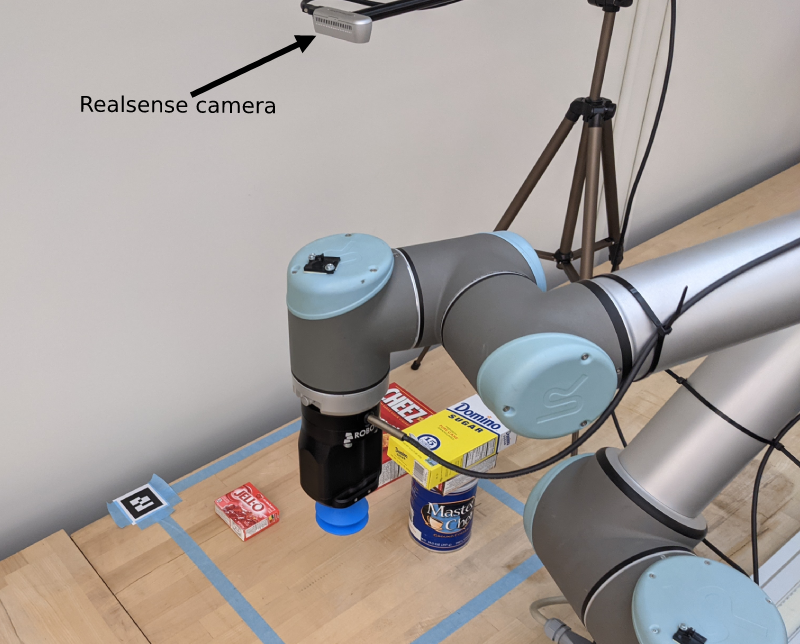}
    % \vspace{-10pt}
    \caption{We use a UR10 robot to execute the manipulation actions predicted by our Scene Exploration Agent (SE$^{*}$) and discover hidden objects. Note that the Realsense camera takes a top-down image of the scene and cannot objects hidden under the sugar box.
    \sehoon{is this mentioned in the main text?}
    }
    \vspace{-4mm}
    \label{fig:real_world}
\end{figure}

\textbf{Real Robot Experiments:} We show evidence of sim-to-real transfer by testing the
Scene Exploration agent (SE) in the real world.
Our hardware experiments consist of a UR10 robot fitted with an e-pick Vacuum gripper and an Intel 435 Realsense camera mounted over the workspace to get a top-down view of the scene (Figure~\ref{fig:real_world}). 
We first process the input image captured by the camera using PoseRBPF~\cite{deng2021poserbpf} to detect the objects in the scene and get their respective poses. We then feed the partial graph $\hat{S}$ to SE and get the manipulation action that can reveal new objects. We repeat this process until all objects have been detected. We show the robot successfully solving $3$ different cluttered scenes (Figure \ref{fig:real_exploration}) in the supplementary video.

%===============================================================================

% \vspace{-2 mm}
\section{Conclusion}
\label{sec:conclusion}
% \vspace{-2 mm}
We investigate generating and interactively exploring structured clutter with multiple hidden objects. We present a graph-based framework to represent, generate and explore complex structured clutter scenes, while ensuring physical stability. We introduce two agents, a Scene Generation agent that skillfully hides objects and a Scene Exploration agent that rearranges the scene to find hidden objects, and demonstrated their effectiveness against multiple baselines.
We showed that our agents generalize to scene sizes not encountered during training and provided additional experimental verification proving the scalability of our method. Finally, we test our Scene Exploration Agent with real world clutter, demonstrating successful sim-to-real transfer. For future work, we intend to test the scalability of our approach to a wider range of objects, in other cluttered scenarios~\cite{savva2019habitat}.
% We modeled the scene with scene graphs that can be used to generate scenes with a system of scene grammar, and that can serve as the backbone of an interactive exploration pipeline. We learned a policy to output production rule sequences for this grammar that generate stable and difficult scenes with multiple hidden objects. We then developed a Scene Exploration Agent that uses learned priors about the physical plausibility of scenes, to rearrange and discover hidden objects. We evaluated the performance of our agents by comparing them to a variety of baselines and showed a significant margin of improvement. 
% \deleted{Our scene grammar can be extended to an arbitrary number of objects, by adding a production rule to insert every new object type.} 

\section*{ACKNOWLEDGMENT}

We thank Dr. Matthew Gombolay for lending us the UR10 robot. This work was in part funded by a grant from Cisco Corporation.

% In the future, we plan on extending our work to generating object arrangements and exploration policies in large scale indoor environments such as in iGibson~\citep{shen2020igibson} and Habitat~\citep{savva2019habitat}. 

% \niranjan{We want to add more objects - 1 production rule per object}
%===============================================================================

% The maximum paper length is 8 pages excluding references and acknowledgements, and 10 pages including references and acknowledgements

\clearpage
% \newpage
% The acknowledgments are automatically included only in the final and preprint versions of the paper.
% \acknowledgments{If a paper is accepted, the final camera-ready version will (and probably should) include acknowledgments. All acknowledgments go at the end of the paper, including thanks to reviewers who gave useful comments, to colleagues who contributed to the ideas, and to funding agencies and corporate sponsors that provided financial support.}

%===============================================================================

\bibliographystyle{IEEEtran}
\bibliography{IEEEabrv,main}
% \newpage
% \appendix
% \newpage
% \input{appendix}
\end{document}